\documentclass[runningheads]{llncs}
\usepackage[utf8]{inputenc}
\usepackage{graphicx}
\usepackage{booktabs} 
\usepackage{amsmath,amssymb,amsfonts}
\usepackage[noend, algoruled, lined, linesnumbered]{algorithm2e}
\usepackage{algcompatible}
\usepackage{caption}
\newsavebox\CBox

\usepackage{textcomp}
\usepackage{hyperref}
\usepackage{comment}
\usepackage{longtable}
\usepackage{multirow}
\usepackage{placeins}
\usepackage{multicol}
\usepackage{subfig}
\usepackage{wasysym}
\usepackage{enumitem}
\usepackage{cite}

\usepackage{cuted}
\usepackage{color,soul}
\usepackage{soulutf8}

\usepackage{comment}

\begin{document}
\title{Ants can orienteer a thief in their robbery}
%
%
\author{Jonatas B. C. Chagas\inst{1,2} 
\and
Markus Wagner\inst{3} 
}
\authorrunning{J. B. C. Chagas and M. Wagner}
%
\institute{Dep. de Computa\c{c}\~{a}o, Universidade Federal de Ouro Preto, Ouro Preto, Brazil \and
Dep. de Inform\'{a}tica, Universidade Federal de Vi\c{c}osa, Vi\c{c}osa, Brazil
\email{jonatas.chagas@\{iceb.ufop.br, ufv.br\}}\\
\and
School of Computer Science, The University of Adelaide, Adelaide, Australia \\
\email{markus.wagner@adelaide.edu.au}}
\maketitle              
\begin{abstract}
The Thief Orienteering Problem (ThOP) is a multi-component problem that combines features of two classic combinatorial optimization problems: Orienteering Problem and Knapsack Problem. The ThOP is challenging due to the given time constraint and the interaction between its components. We propose an Ant Colony Optimization algorithm together with a new packing heuristic to deal individually and interactively with problem components. Our approach outperforms existing work on more than 90\% of the benchmarking instances, with an average improvement of over 300\%.
\end{abstract}

\section{Introduction}
\label{sec:introduction}

The Traveling Thief Problem (TTP)~\cite{bonyadi2013travelling} is a well-studied multi-component problem that combines Traveling Salesman Problem (TSP) and Knapsack Problem (KP). The TTP has been designed in order to provide an academic abstraction of multi-component problems for the scientific community. In brief, in the TTP, a single thief has to visit all cities (TSP component) and can make a profit by stealing items and storing them in a rented knapsack (KP component). As stolen items are stored in the knapsack, it becomes heavier, and the thief travels more slowly, with a velocity inversely proportional to the knapsack weight. The thief's objective is to maximize the total profit of the stolen items while considering the price to pay for the knapsack, which is proportional to the rent time.

The TTP has been gaining fast attention due to its challenging interconnected multi-components structure. Thus far, many approaches have been proposed for solving it, including iterative heuristics~\cite{polyakovskiy2014comprehensive}, metaheuristic approaches~\cite{el2015cosolver2b, faulkner2015approximate, el2016population, wagner2016stealing}, and exact approaches to study the quality of solutions for small instances~\cite{wu2017exact}. Some studies have investigated the structure and properties of the TTP~\cite{mei2016investigation,yafrani2018fitness}. We refer~\cite{wagner2018case} for a comparison of 21 algorithms in order to provide a TTP portfolio.

The Thief Orienteering Problem (ThOP)~\cite{santos2018thief} has been designed as an academic multi-component problem with different interactions and constraints in mind: it combines the Orienteering Problem (OP) and the Knapsack Problem (KP). The OP is a well-studied problem in operational research (see, e.g., \cite{GoLeVo87, chao1996fast, vansteenwegen2011orienteering,GUNAWAN2016315}), where a participant starts on a given point, travels through a region visiting checkpoints, and has to arrive at a control point within a given time. Each checkpoint has a score, and the objective of the participant is to find the route that maximizes the total score, i.e., whose sum of scores of the checkpoints visited is maximal. Recent real-world examples of the OP include tourists planning their sight-seeing trips~\cite{Fang2014travelrouting}, rescue teams planning the visit of safe places in case of emergencies~\cite{Baffo2017emergencyrouting}, and politicians or music bands planning their tours~\cite{AksenS2016politicianrouting,Freeman2018bandrouting}.

For the ThOP, Santos and Chagas~\cite{santos2018thief} have proposed a Mixed Integer Non-Linear Programming formulation for it, but no computational results have been presented due to the formulation's complexity. Instead, two simple heuristic algorithms have been proposed, i.e., one based on Iterated Local Search (ILS)~\cite{lourencco2003iterated} and one based on a Biased Random-Key Genetic Algorithm (BRKGA)~\cite{gonccalves2011biased}. The BRKGA outperformed ILS on large instances, and the authors have attributed this to the diversification introduced of the mutant individuals.

In this work, we propose the use of a two-phase swarm intelligence approach based on Ant Colony Optimization (ACO) and a new greedy heuristic, to construct, respectively, the route and the packing plan (stolen items) of the thief. We investigate the importance of the components via automated algorithm configuration and then evaluate our approach on a broad set of instances. 

The remainder of this paper is structured as follows. In Section~\ref{sec:problem_description}, we formally describe the ThOP and present detailed solution examples to demonstrate the interwovenness characteristic of the multi-components of the problem. In Section~\ref{sec:max_min_ant_algorithm}, we present our solution approach proposed for addressing the ThOP. Section~\ref{sec:computational_experiments} reports the experiments and analyzes the performance of the proposed solution approach against previous ones already proposed in the literature. We conclude in  Section~\ref{sec:conclusions} with a summary and outline possible future work. 

\section{Problem description}
\label{sec:problem_description}

As stated by Santos and Chagas~\cite{santos2018thief}, in the ThOP, there is a set of $n$ cities, labeled from $1$ to $n$, where the cities $1$ and $n$ are, respectively, the cities where the thief starts and ends their journey. A set of $m$ items scattered among the other cities $(2, \ldots, n-1)$, and each city has one or more items. Each item $i \in \{1, \ldots, m\}$ has a profit $p_i$ and weight $w_i$ associated. For any pair of cities $i$ and $j$, the distance $d_{ij}$ between them is known. 

In the ThOP, there is a single thief to steal items scattered among cities. The thief has a knapsack with a limited capacity $W$ to carry the items. Moreover, the thief has a maximum time $T$ to complete their whole robbery plan. The speed of the thief is inversely proportional to their knapsack weight. When the knapsack is empty, the thief can move with their maximum speed $v_{max}$. However, when the knapsack is full, the thief moves with the minimum speed $v_{min}$. The speed $v$ of the thief when the knapsack weight is $w \leq W$ is given by $v = v_{max} - w \times (v_{max} - v_{min})\,/\, W$.

The objective of the ThOP is to provide a path from the start city $1$ to the end city $n$, as well as a set of items chosen from the cities visited throughout the route so that to maximize the total profit stolen, ensuring that the capacity of the knapsack $W$ is not surpassed and the total traveling time of the thief is within the time limit $T$. The thief does not need to visit all cities.

Note that, while the ThOP and the TTP appear to be similar due to the KP as a component, it has been argued that the ThOP is more practical due to two key differences: in the ThOP there is (A) no need to visit all the cities, and (B) the interaction is not through a time-dependent rent for the knapsack, but through a constraint that imposes on the thief a time limit to complete the tour -- at the very least for the aforementioned real-world examples, speed typically remains constant, but time constraints have to be fulfilled, and only worthwhile places have to be visited. 
While the relaxation of difference (A) might appear trivial, the consideration of this constraint, i.e., to visit all cities, is typically reflected in the design of heuristic~\cite{wagner2018case} and exact~\cite{wu2017exact,Neumann2019fptasPWT} approaches to the TTP, with Chand and Wagner~\cite{chand2016fast}'s Multiple Traveling Thieves Problem (MTTP) being the only exception known to us. Regarding the difference (B) and the ThOP in general, applications of it can arise when there is no enough time and capacity to visit all possible cities. For an overview of time-dependent routing problems, we refer the interested reader to Gendreau et al.~\cite{gendreau2015time}.

\begin{figure}[!ht]
	\centering
	\includegraphics[scale=0.40]{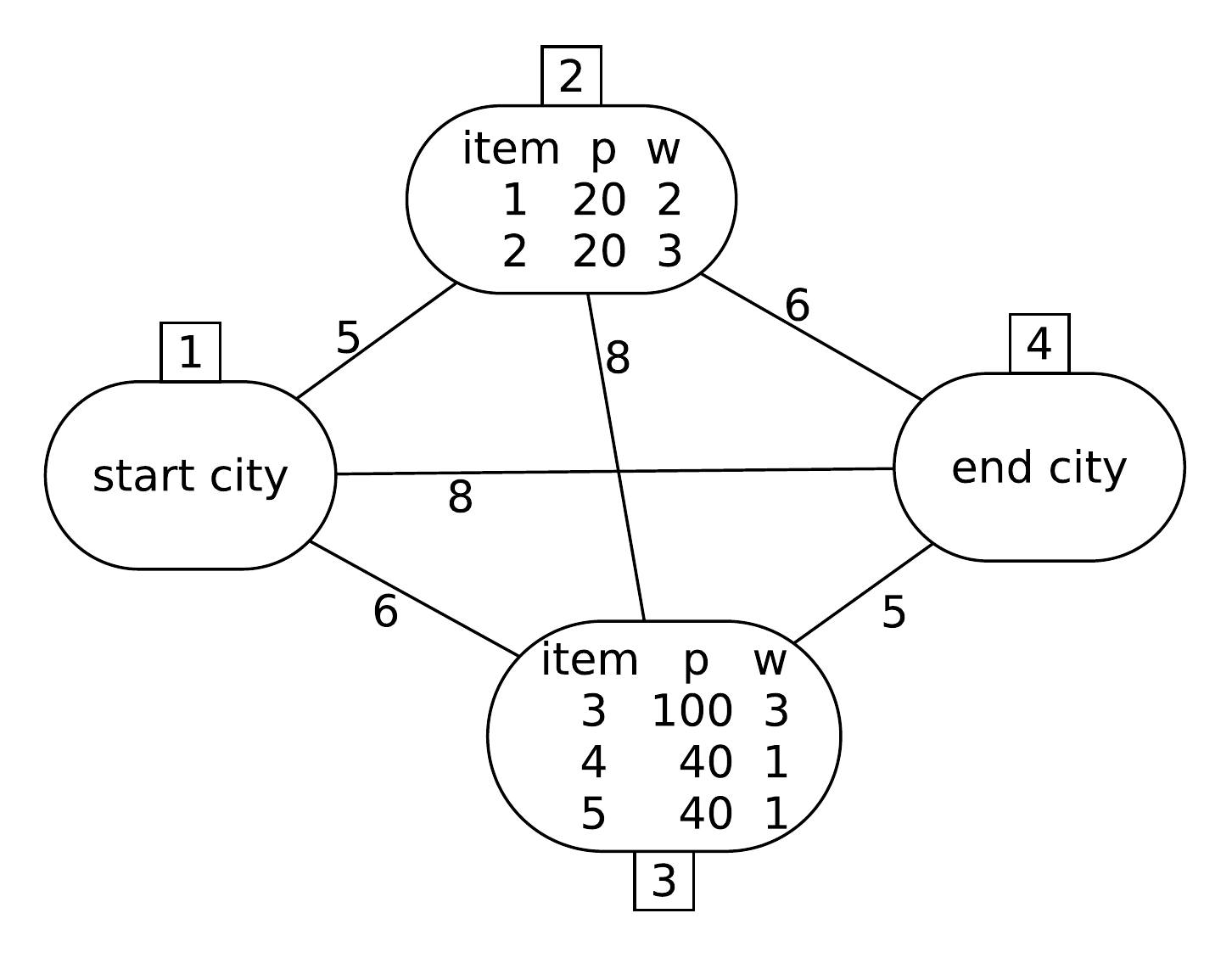}\vspace{-3mm}
	\caption{A ThOP instance involving 4 cities and 5 items (from \cite{santos2018thief}).}
	\label{fig:example}
\end{figure}

In order to clarify the characteristics of the ThOP, we depict in Figure~\ref{fig:example} a small worked example of a ThOP instance that involves 4 cities and 5 items. Note that there are no items in the start (1) and end (4) cities, whereas there are some items of different weights and profit distributed in the other cities (2 and 3). The distances from each pair of cities are given in the edges. In the following, we present in detail some solutions for this instance. For this purpose, let us consider $v_{min} = 0.1$, $v_{max} = 1.0$, $W = 3$, and $T = 75$.

We may represent a ThOP solution in two parts $(\pi, z)$. The first one consists of the route $\pi = \langle 1, \ldots, n \rangle$, a vector containing the ordered list of visited cities. Note that the first and last city are fixed for any feasible solution. The second part is the packing plan $z = \langle z_1, z_2, \ldots, z_m \rangle$, a binary vector representing the states of items ($z_i = 1$ if item $i$ is stolen, and $0$ otherwise). According to this representation, let us consider the following ThOP solutions for the instance previously described:

\begin{itemize}[leftmargin=5mm]
    \item {
        $(\langle 1, 2, 3, 4 \rangle, \langle 1, 0, 0, 1, 0 \rangle)$: it is a feasible solution with a total profit of $20 + 40 = 60$. The total weight of stolen items is $3$ and the total traveling time is $75$, which satisfies both limits $W$ and $T$. The total traveling time is calculated as:
        \begin{itemize}
            \setlength\itemsep{0mm}
            \item travel from the start city to city $2$ at maximum speed: time is computed as $d_{12}/v_{max} = 5/1.0 = 5$;
            \item at city $2$ the thief steals item $1$: the speed decreases to $v = 1.0 - 2 \times (1.0 - 0.1)\,/\,3 = 0.4$;
            \item travel from city $2$ to city $3$: total traveling time is $5 + d_{23}/v = 5 + 8/0.4 = 5 + 20 = 25$;
            \item at city $3$ item $4$ is collected: the speed drops to $v = 1.0 - 3 \times (1.0 - 0.1)\,/\,3 = 0.1$;
            \item travel from city $3$ to the end city: total traveling time is $5 + 20 + d_{34}/v = 5 + 20 + 5/0.1 = 5 + 20 + 50 = 75$.
        \end{itemize}
    }
    \item {
        $(\langle 1, 3, 2, 4 \rangle, \langle 1, 0, 0, 1, 0 \rangle)$: it is an infeasible solution. Although the stolen items have been the same as the previous solution, the total traveling time $(83.43)$ exceeds the time limit: 
        \begin{itemize}
            \setlength\itemsep{0mm}
            \item travel from the start city to city $3$ at maximum speed: time is computed as $d_{13}/v_{max} = 6/1.0 = 6$;
            \item at city $3$ the thief steals item $4$: the speed decreases to $v = 1.0 - 1 \times (1.0 - 0.1)\,/\,3 = 0.7$;
            \item travel from city $3$ to city $2$: total traveling time is $6 + d_{32}/v = 6 + 8/0.7 = 6 + 11.43 = 17.43$;
            \item at city $3$ item $4$ is collected: the speed drops to $v = 1.0 - 3 \times (1.0 - 0.1)\,/\,3 = 0.1$;
            \item travel from city $2$ to the end city: total traveling time is $6 + 17.43 + d_{24}/v = 6 + 17.43 + 6/0.1 = 6 + 17.43 + 60 = 83.43$.
        \end{itemize}
    }
    \item {
        $(\langle 1, 3, 4 \rangle, \langle 0, 0, 1, 0, 0 \rangle)$: it is the optimal solution for this instance with a total profit of $100$. The total weight is $3 \leq W$ and the total traveling time is $56 \leq T$: 
        \begin{itemize}
            \setlength\itemsep{0mm}
            \item travel from the start city to city $3$ at maximum speed: time is computed as $d_{13}/v_{max} = 6/1.0 = 6$;
            \item at city $3$ the thief steals item $3$: the speed decreases to $v = 1.0 - 3 \times (1.0 - 0.1)\,/\,3 = 0.1$;
            \item travel from city $3$ to the end city: total traveling time is $6 + d_{34}/v = 6 + 5/0.1 = 6 + 50 = 56$.
        \end{itemize}
    }
\end{itemize}

Note that the packing plan of the optimal ThOP solution for the example instance happens to be the same as the optimal solution for the knapsack problem. However, it is not always that the thief can steal the best knapsack configuration within the time limit $T$. To exemplify this, let us now consider a tighter time limit equal to 20 for the previous instance. For this case, the optimal ThOP solution would be $(\langle 1, 3, 4 \rangle, \langle 0, 0, 0, 1, 1 \rangle)$, which has a total profit of $80$ and total traveling time of $18.5$.

\section{Stealing items with ants}
\label{sec:max_min_ant_algorithm}

In the following, we describe our heuristic approach for the ThOP. It is loosely based on Wagner's TTP study~\cite{wagner2016stealing}. As in~\cite{wagner2016stealing}, we propose in this work the use of swarm intelligence based on Ant Colony Optimization (ACO)~\cite{dorigo1999ant} in order to solve ThOP's tour part, while a novel heuristic will be responsible for solving the ThOP's packing part, i.e., to select the set of stolen items. 

ACO algorithms consist of an essential class of probabilistic search techniques that are inspired by the behavior of real ants. These algorithms have proven to be efficient in solving a range of combinatorial problems \cite{dorigo2005ant}. The basic idea behind ACOs is that ants construct solutions for a given problem by carrying out walks on a so-called construction graph. These walks are influenced by the pheromone values that are stored along the edges of the graph. During the optimization process, the pheromone values are updated according to good solutions found during the optimization, which should then lead the ants to better solutions in further iterations of the algorithm. We refer the interested reader to the book by Dorigo and Birattari~\cite{dorigo2010ant} for a comprehensive introduction.

In order to define the thief's route, we use Stützle's ACOTSP 1.0.3 framework\footnote{Publicly available online at \href{http://www.aco-metaheuristic.org/aco-code}{\textcolor{blue}{http://www.aco-metaheuristic.org/aco-code}}}. This framework implements several ACO algorithms for the symmetric TSP, i.e., its found solutions are tours that visit all cities. While this may not be efficient or even feasible for the thief due to the time limit to conclude their journey, we will adapt the output according to the solution found by the packing algorithm in order to determine efficient solutions for the ThOP.

We note that ACOTSP builds complete TSP tours, not OP tours, hence possibly affecting the algorithm performance. We decided against the OP-tour approach: assuming that we have an OP tour and then consider the packing, we may (based on our two-phase approach) end up skipping cities anyhow if there are no interesting items to pick up; hence, a further dropping of cities may be required anyhow. Of course, this would be different if we would have an algorithm to solve the OP and KP parts simultaneously.

\subsection{ACO framework and adjustments}

The ACOTSP framework allows us to choose which ant colony optimization approach to be used. As in~\cite{wagner2016stealing}, we use the standard MAX-MIN ant system by Stützle and Hoos~\cite{stutzle2000max}, which restricts all pheromones to a bounded interval in order to prevent pheromones from dropping to arbitrarily small values. In Algorithm~\ref{alg:acothop}, we show the simplified overview of the proposed swarm intelligence approach, combined with the packing heuristic algorithm. 

\begin{algorithm}[!ht]
\makeatletter
\newcommand{\algorithmfootnote}[2][\footnotesize]{%
  \let\old@algocf@finish\@algocf@finish
  \def\@algocf@finish{\old@algocf@finish
    \leavevmode\rlap{\begin{minipage}{\linewidth}
    #1#2
    \end{minipage}}%
  }%
}
\footnotesize
\DontPrintSemicolon
\SetKwData{Left}{left}
\SetKwData{Up}{up}
\SetKwFunction{FindCompress}{FindCompress}
\SetKwInOut{Input}{input}
\SetKwInOut{Output}{output}
$\pi^{best} \gets \varnothing, z^{best} \gets \varnothing$ \label{alg:best_sol_init} \\
\Repeat{\upshape stopping condition is fulfilled} { \label{alg:stopping_criterion_begin}
    $\Pi \gets$ construct TSP tours using ants \label{alg:construct_tsp_tours} \\
    \ForEach{\upshape TSP tour $\pi \in \Pi$} { \label{alg:for_each_tour_begin}
        $z \gets$ construct a packing plan from $\pi$ using \textsc{Pack($\pi$,~$ptries$)} \label{alg:packing_plan} \\
        \If{\upshape profit of $z$ is higher than profit of $z^{best}$} { \label{alg:second_replace_begin}
            $\pi^{best} \gets \zeta(\pi$), $z^{best} \gets z$ \label{alg:clear_tsp_tour}
        } \label{alg:second_replace_end}
    } \label{alg:for_each_tour_end}
    update ACO statistics and pheromone trail \label{alg:update_pheromone} \\
} \label{alg:stopping_criterion_end}
\Return $\pi^{best}$, $z^{best}$
\caption{ACO for the ThOP}
\label{alg:acothop}
\algorithmfootnote{$\zeta(\pi)$ returns a tour for the thief from the TSP tour $\pi$ by removing all cities where no item is stolen according to the packing plan $z$.}
\end{algorithm}%

Initially (line~\ref{alg:best_sol_init}), the best ThOP solution (tour and packing plan) found by the algorithm is initialized as an empty solution. The algorithm performs its iterative cycle (lines~\ref{alg:stopping_criterion_begin} to \ref{alg:stopping_criterion_end}) as long as the stopping criterion is not fulfilled. At line~\ref{alg:construct_tsp_tours}, each ant constructs a TSP tour. For each TSP tour $\pi$ (line~\ref{alg:for_each_tour_begin}), we apply our heuristic algorithm for defining a packing plan (line~\ref{alg:packing_plan}, Algorithm~\ref{alg:packing_algorithm}), thus defining a feasible ThOP solution $(\pi, z)$. At lines~\ref{alg:second_replace_begin} to \ref{alg:second_replace_end}, the best solution found is possibly updated according to the solution $(\pi, z)$ previously found. Note that we remove from $\pi$ all cities where no items have been stolen according to the packing plan $z$ (line \ref{alg:clear_tsp_tour}) in order to get a more efficient ThOP's tour (all ThOP instances use Euclidean distances rounded up). After all tours have been considered, ACO statistics and the pheromone values are updated according to the quality of the ThOP solutions found (line \ref{alg:update_pheromone}). At the end of the algorithm (line~\ref{alg:return_best_packing_plan}), the best solution found is returned.

\paragraph{Implementation Notes} The overall logic of the ACOTSP framework remains unchanged in our proposed algorithm. Some minimal modifications have been performed to adapt it to the ThOP specifications. To construct the TSP tours, we just established that the first and last cities must be those where the thief begins and ends their robbery journey. In the ACOTSP framework, the pheromone trail update performs based on the quality of the TSP tours found by ants. Since the objective of the TSP is to find the shortest possible tour visiting each city, the fitness of a given tour is inversely proportional to its total distance. On the other hand, in our ACOTSP adaptation, the fitness of each tour is set in terms of the quality of the stolen items throughout the tour, which are defined by the heuristic packing plan. As the ACOTSP framework is developed explicitly for the TSP, a minimization problem where its solutions have positive objective values, we consider that the fitness of a ThOP's tour $\pi$ is inversely proportional to $UB + 1 - p(z)$, where $UB$ is an upper bound for the ThOP and $p(z)$ is the total profit of packing plan $z$. Note that in this way we can maintain the same behavior of fitness of the TSP solutions, without modifying the ACO framework structure. The upper bound $UB$ is defined as the optimal solution for the KP version that allows selecting fractions of items. This KP version can be solved in $O(m\log_2 m)$.

\subsection{ThOP packing heuristic}

In Algorithm \ref{alg:packing_algorithm}, we describe our heuristic strategy for constructing a packing plan from a fixed tour. Note that even when the tour of the thief is kept fixed, finding the optimal packing configuration is NP-hard~\cite{polyakovskiy2015packing}.

\begin{algorithm}[!ht]
\footnotesize
\DontPrintSemicolon
\SetKwData{Left}{left}
\SetKwData{Up}{up}
\SetKwFunction{FindCompress}{FindCompress}
\SetKwInOut{Input}{input}
\SetKwInOut{Output}{output}
$z \gets \varnothing$, $try \gets 1$ \\
\Repeat{$try > ptries$} { \label{alg:repeat_begin}
    choose a real number for each parameter $\theta$, $\delta$, and $\gamma$ from a uniform distribution in the range [0, 1], so that $\theta + \delta + \gamma = 1$ \label{alg:random_values} \\
    \ForEach{\upshape $i \gets 1$ \textbf{to} $m$} { \label{alg:compute_scores_begin}
        compute score $s_i$ for item $i$ \tcp*{Eq. \ref{eq:score}} 
    } \label{alg:compute_scores_end}
    $z' \gets \varnothing$ \\
    \For{\upshape $j \gets 1$ \textbf{to} $m$} { \label{alg:packing_begin}
        $i \gets $ get item with the $j$-th highest score \\
        $z' \gets z' \cup \{i\}$ \\
        \lIf{\upshape weight of $z'$ is higher than $W$} { \label{alg:weight_constraint}
            $z' \gets z' \setminus \{i\}$ 
        }
        \Else{
            $t \gets $ compute the required time to steal $z'$ by visiting only cities with selected items following the order of the TSP tour $\pi$ \label{alg:calculate_time} \\
            \lIf{\upshape $t$ is longer than $T$} { \label{alg:time_constraint}
                $z' \gets z' \setminus \{i\}$  
            }
        }
    } \label{alg:packing_end}
    \lIf{\upshape profit of $z'$ is higher than profit of $z$} { \label{alg:best_packing_plan}
        $z \gets z'$ \label{alg:update_best_packing_plan}
    }
    $try \gets try + 1$ \\
} \label{alg:repeat_end}
\Return $z$ \label{alg:return_best_packing_plan}
\caption{Packing Algorithm: \textsc{Pack($\pi$, $ptries$)}}
\label{alg:packing_algorithm}
\end{algorithm}%

Our packing heuristic algorithm seeks to find a good packing plan $z$ from multiple attempts for the same tour $\pi$. The number of attempts is defined by $ptries$. Each attempt is described between lines~\ref{alg:repeat_begin} to \ref{alg:repeat_end}. At the beginning of each attempt (line~\ref{alg:random_values}), we uniformly select three random values ($\theta$, $\delta$, and $\gamma$) between 0 and 1, and then normalize them so that their sum is equal to 1. These values are used to compute a score $s_i$ for each item $i \in \{1, \ldots, m\}$ (lines~\ref{alg:compute_scores_begin} to \ref{alg:compute_scores_end}), where $\theta$, $\delta$, and $\gamma$ define, respectively, exponents applied to profit $p_{i}$, weight $w_i$, and distance $d_{i}$ in order to manage their impact. The distance $d_{i}$ is calculated according to the tour $\pi$ by sum all distances from the city where is the item $i$ to the end city. Equation~\ref{eq:score} shows as the score of item $i$ is calculated.

\begin{equation}
    \label{eq:score}
    s_{i} = \frac{{p_{i}}^{\theta}}{{w_{i}}^{\delta} \times {d_{i}}^{\gamma}}
\end{equation}

Note that each score $s_i$ incorporates a trade-off between a distance that item $i$ has to be carried over, its weight, and its profit. Equation \ref{eq:score} is based on the heuristic \textsc{PackIterative} that has been developed for the TTP~\cite{faulkner2015approximate}. However, unlike in~\cite{faulkner2015approximate}, we consider an exponent for the term of distance to vary the importance of its influence. Furthermore, the values of all exponents are randomly selected drawn between 0 and 1 for each attempt (and then normalized) to search the space for greedy packing plans.

After computing scores for all items, we use their values to define the priority of each item in the packing strategy. The higher the score of an item, the higher its priority. Between lines~\ref{alg:packing_begin} and \ref{alg:packing_end}, we create the packing plan for the current attempt by considering the items according to their priorities. If an item violates the constraints of the ThOP (lines~\ref{alg:weight_constraint} and \ref{alg:time_constraint}), it is not selected. Note that we calculate travel time (line~\ref{alg:calculate_time}) from the cities listed on tour $\pi$, but we ignore those cities where no items are selected. After completing the current attempt's packing plan, its quality is compared to the best packing plan so far (line~\ref{alg:best_packing_plan}), which is then possibly updated (line~\ref{alg:update_best_packing_plan}). At the end of all attempts, the best packing plan found is returned (line~\ref{alg:return_best_packing_plan}).

Note that our packing algorithm is non-deterministic (in contrast to the deterministic \textsc{PackIterative}~\cite{faulkner2015approximate}'s), as it has randomized components. In our preliminary experiments, we have observed that ants find identical or very similar routes throughout the iterations of the ACO algorithm. For this reason, we decided to design our packing algorithm in a non-deterministic way in order to increase the explore the packing plan space more broadly. Moreover, via the parameter $ptries$, we can control the number of attempts needed to reached good packing plans.

\section{Computational experiments}
\label{sec:computational_experiments}

We now present the experiments performed to study the performance of the proposed framework concerning the quality of its solutions. We have rerun Santos and Chagas~\cite{santos2018thief}'s ThOP code to enable a fair comparison as the computational budget is based on wallclock time.

Our framework has been implemented based on Thomas Stützle’s ACOTSP 1.0.3 framework, which has been implemented in C programming language. In our experiments, each run of the proposed algorithm has been sequentially (nonparallel) performed on an Intel(R) Xeon(R) E5-2660 (2.20GHz), running under CentOS Linux 7 (Core). Our code, as well as all results and solutions, can be found at 
\href{https://github.com/jonatasbcchagas/aco_thop}{\textcolor{blue}{https://github.com/jonatasbcchagas/aco\_thop}}. 

\subsection{Benchmarking instances}

To assess the quality of the proposed algorithm, we have used all ThOP instances defined by Santos and Chagas~\cite{santos2018thief}. As stated by the authors, these instances have been created upon a benchmark of TTP instances~\cite{polyakovskiy2014comprehensive} by removing the items on city $n$ and by adding a maximum travel time. They have created 432 instances with the following characteristics:

\begin{itemize}
    \item {
        numbers of cities: 51, 107, 280, and 1000
        (TSP instances ({\tt XXX}): {\it eil51}, {\it pr107}, {\it a280}, {\it dsj1000});
    }
    \item {
        numbers of items per city ({\tt YY}): {\it 01}, {\it 03}, {\it 05}, and {\it 10};
    }
    \item {
        types of knapsacks ({\tt ZZZ}): weights and values of the items are bounded and strongly correlated ({\it bsc}), uncorrelated ({\it unc}), or uncorrelated with similar weights ({\it usw});
    }
    \item {
        sizes of knapsacks ({\tt WW}): {\it 01}, {\it 05} and {\it 10} times the size of the smallest knapsack;
    }
    \item {
        maximum travel times ({\tt TT}): {\it 01}, {\it 02}, and {\it 03} classes. These values refer to 50\%, 75\%, and 100\% of instance-specific references times defined in the original ThOP paper~\cite{santos2018thief}.
    }
\end{itemize}

All 432 ThOP instances can be obtained by combining the different characteristics described above. Each instance is identified as {\tt XXX\_YY\_ZZZ\_WW\_TT.thop}.

\subsection{Parameter tuning to gain insights}
\label{sec:parameter_tuning}

Our first study analyzes the influence of the values of the main parameters of our algorithm. As in the previous work~\cite{santos2018thief} on the ThOP, we have defined as stopping criteria the execution time equal to $\lceil\frac{m}{10}\rceil$ seconds, which is given in terms of the number of items $m$ of each particular instance. 

The ACOTSP framework allows setting a large number of parameters. We consider the following: \textit{ants} defines the number of ants used; \textit{alpha} controls the relative importance of pheromone trails in the construction of tours; \textit{beta} defines the influence of distances between cities for construction the tours; and \textit{rho} sets the evaporation rate of the pheromone trail. Besides, we analyze the influence of our parameter \textit{ptries}, which is used for deciding how many attempts our packing algorithm performs to determine the set of stolen items.

Table \ref{table:parameter_values} shows the parameter values we have considered in our analysis. The ranges have been selected following preliminary experiments.

\begin{table}[!ht]
\centering
\footnotesize
\centering
\caption{Parameter values considered during the tuning experiments.}
\setlength{\tabcolsep}{15pt}
\begin{tabular}{cc}
\toprule
\multicolumn{1}{c}{Parameter} & \multicolumn{1}{c}{Investigated values} \\ 
\midrule
ants & $\{10, 20, 50, 100, 200, 500, 1000\}$ \\ 
alpha & $\{0.00, 0.01, 0.02, \ldots, 10.00\}$ \\ 
beta & $\{0.00, 0.01, 0.02, \ldots, 10.00\}$ \\ 
rho & $\{0.00, 0.01, 0.02, \ldots, 1.00\}$ \\ 
ptries & $\{1, 2, 3, 4, 5\}$ \\
\bottomrule
\end{tabular}
\label{table:parameter_values}
\end{table}

In order to find a suitable configuration of parameters among all possible ones, we use the Irace package \cite{lopez2016irace}, which is an implementation of the method I/F-Race \cite{birattari2010f}. The Irace package implements an iterated racing framework for the automatic configuration of algorithms. In our experiments, we have used all Irace default settings, except for the parameter \textit{maxExperiments}, which has been set to 5000. This parameter defines the stopping criteria of the tuning process. We refer the readers to \cite{lopez2016iraceguide} for a complete user guide of Irace package.

To analyze the influence of parameter values across the different types of instances, we divide all 432 instances into 48 groups and then execute Irace on each of them. Each group is identified as {\tt XXX\_YY\_ZZZ}, where {\tt XXX} informs the TSP base group, {\tt YY} the number of items per city and {\tt ZZZ} the type of knapsack. Each group {\tt XXX\_YY\_ZZZ} contains all nine instances defined with different sizes of knapsacks and maximum travel time.
 
\begin{figure*}[!ht]
    \centering
 	\setcounter{subfigure}{0}
 	\subfloat {
      		\includegraphics[width=0.38\linewidth]{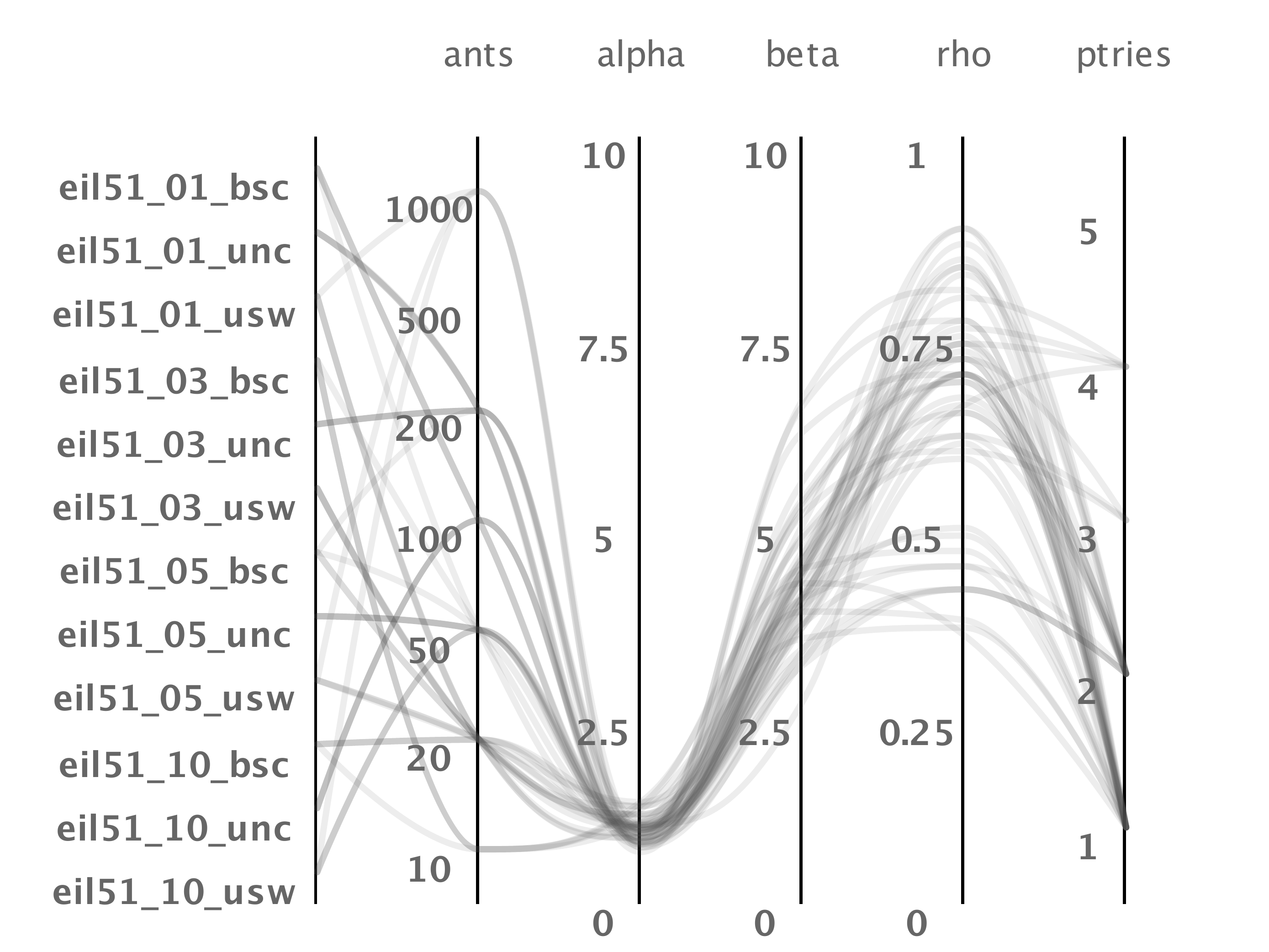}
 	}%
 	\qquad \qquad
 	\subfloat {
      		\includegraphics[width=0.38\linewidth]{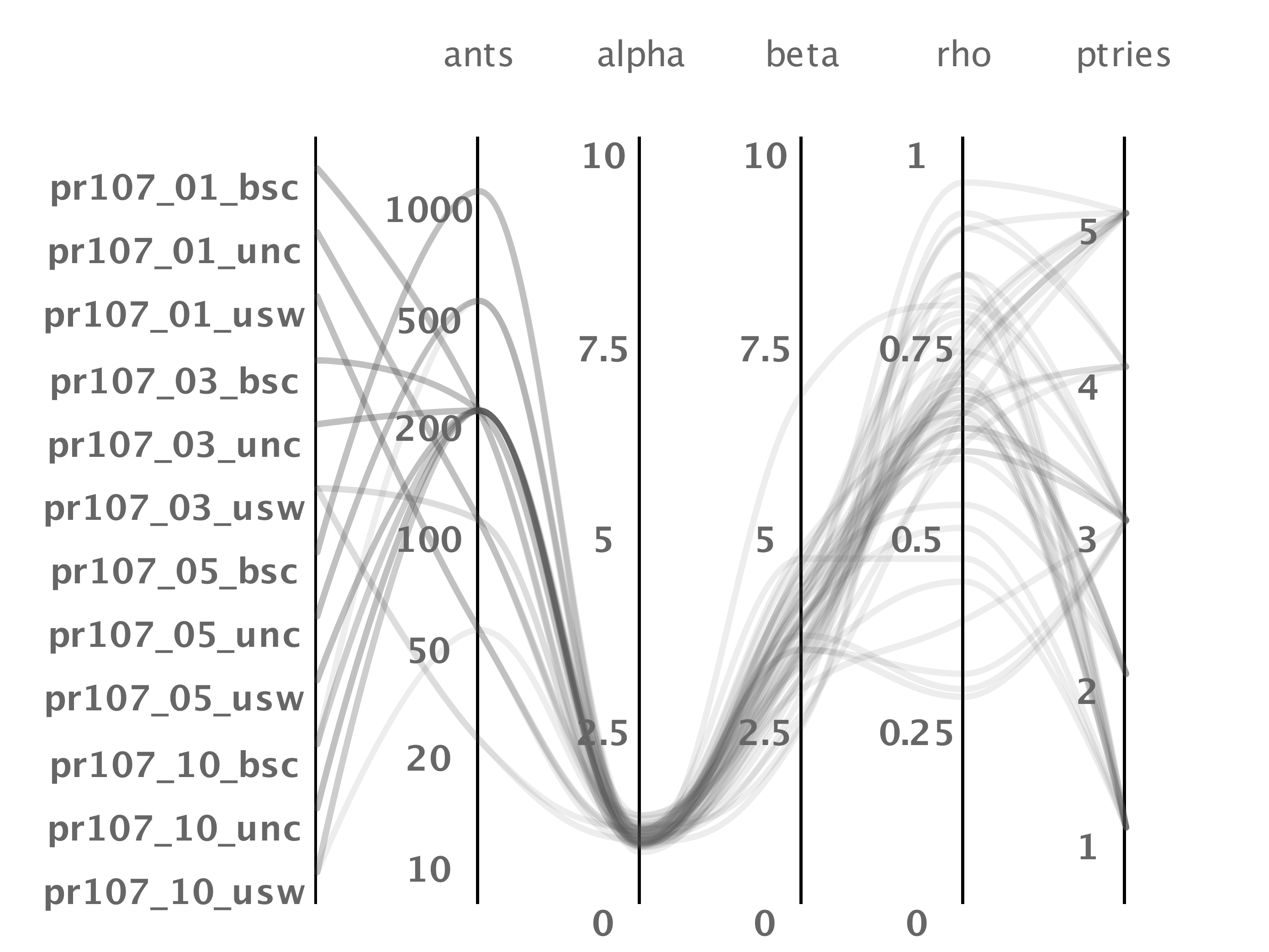}
 	}%
 	
 	\subfloat {
      		\includegraphics[width=0.38\linewidth]{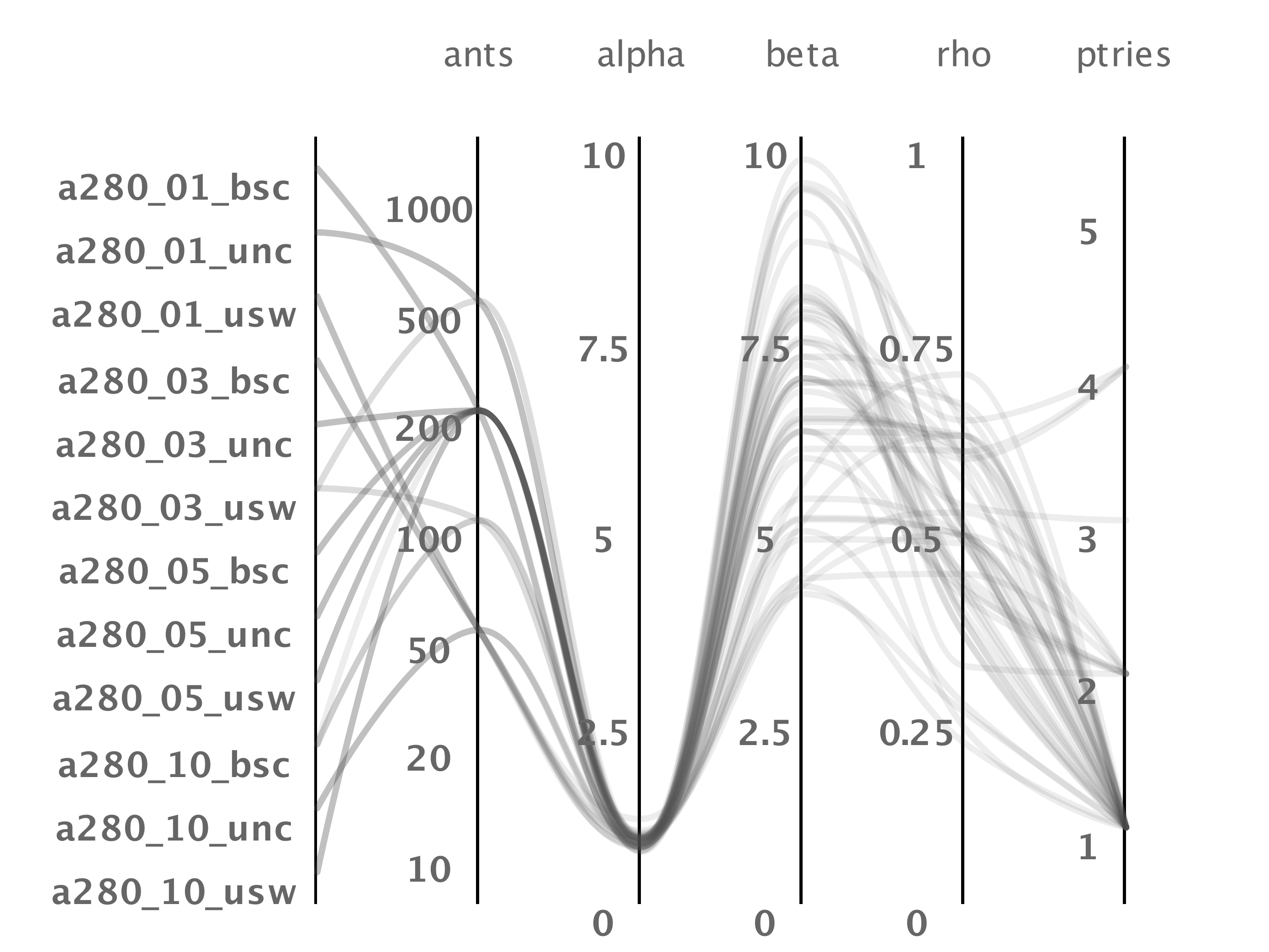}
 	}%
 	\qquad \qquad
 	\subfloat {
      		\includegraphics[width=0.38\linewidth]{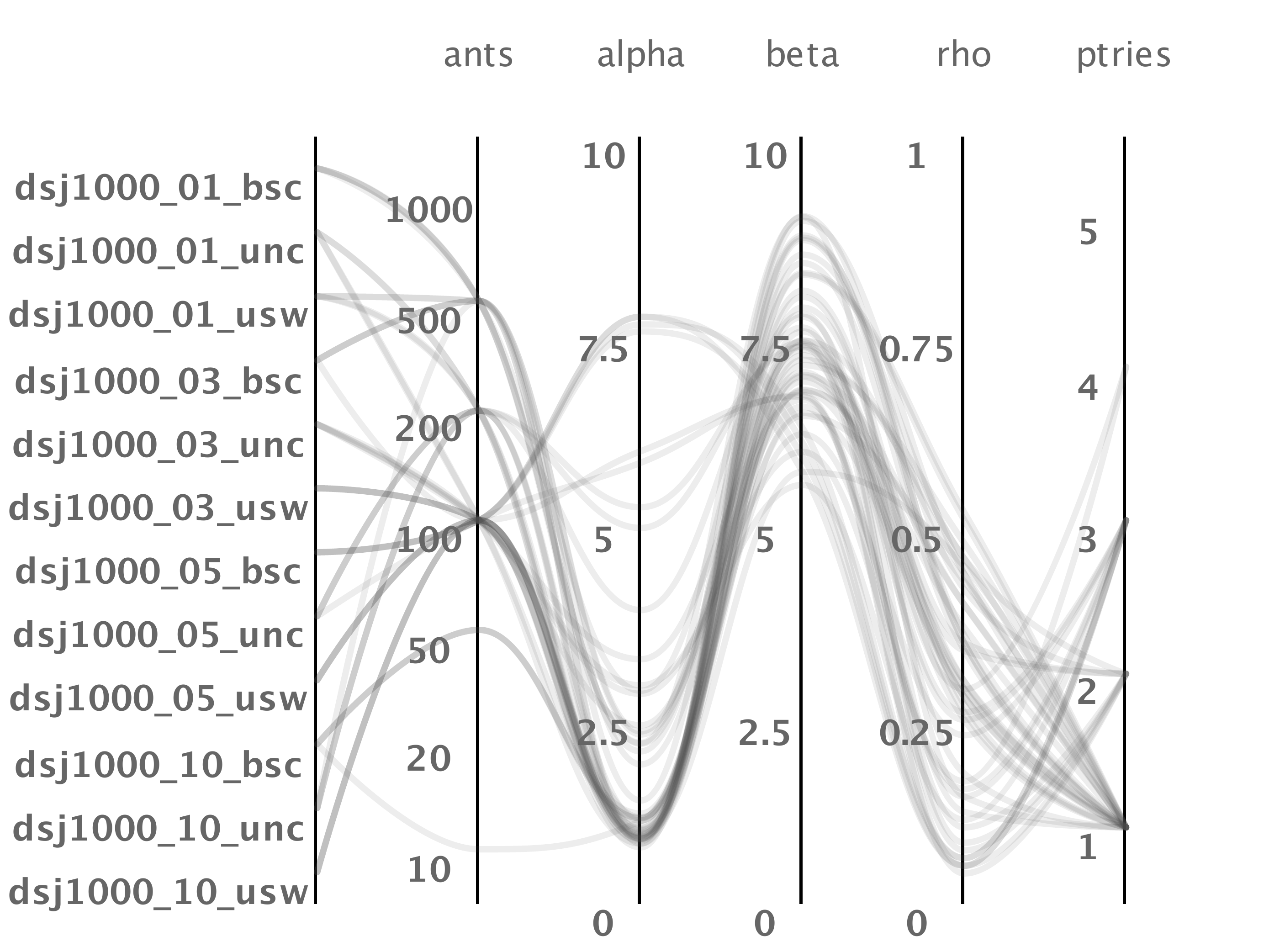}
 	}%
\caption{Best parameter configurations for the 48 groups of instances.}
\centering
\label{fig:irace_results}
\end{figure*}

In Figure \ref{fig:irace_results}, we plot for each group all configurations returned by Irace at the end of its run. Each parallel coordinate plot lists for each of the 48 groups (shown in the left-most column) the configurations returned by Irace (shown in the other columns). As Irace can return more than one configuration, multiple configurations are sometimes shown. Each axis indicates a parameter and its range of values, and each configuration of parameters is described by a line that cuts each parallel axis in its corresponding value. Through the concentration of the lines, we can see which parameter values have been most selected among all tuning experiments. 

We can make several observations. 
For example, the number of ants has a higher concentration between 50 and 200, with a higher frequency between 100 and 200 for the groups of instances that consider the TSP bases {\it pr107}, {\it a280}, and {\it dsj1000}. The importance of the pheromone trail has remained with values close to 1 for all groups of instances. 
This is generally compensated by the values of \textit{beta}, which varies based on the underlying TSP instance. This is not too surprising, as the underlying TSP instances are different in nature and not normalized, hence requiring different values of beta. We can also observe that only few packing attempts (as exhibited by the low \textit{ptries} values) are needed to reach good results, which is especially true for larger instances. 

In an attempt to furnish a single configuration of parameters that can generalize all tuning results and also be able to provide a more appropriate configuration for new unknown instances, we average the numerical values and take the mode of the categorical parameter \textit{ptries}. This results in the following configuration: \textit{ants}~=~196, \textit{alpha}~=~1.24, \textit{beta}~=~5.46, \textit{rho}~=~0.51, and \textit{ptries}~=~1. 

\subsection{Results}
\label{sec:results}

In order to analyze the efficiency of the proposed algorithm on all ThOP instances, we run our ACO algorithm 10 independent times on each instance, and then use the average value of the objective function and the best one found in these runs in our analysis. Our experiments analyze two versions of our ACO algorithm. In the first one, we consider the algorithm set with the best parameter values found by the Irace package (collectively called ACOThOP*). In contrast, the second version uses the general configuration of parameters derived from the Irace results of all tuning experiments (called ACOThOP).

\begin{figure*}[!ht]%
\centering
\includegraphics[width=\linewidth]{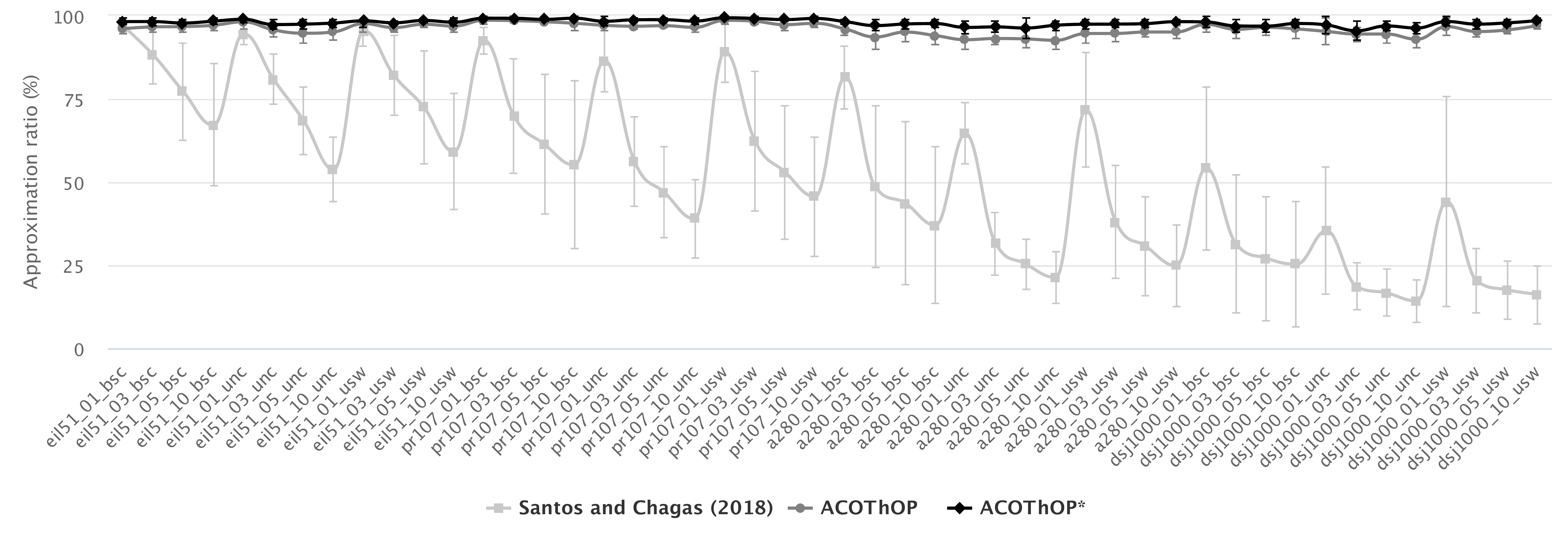}
\caption{Approximation ratio of the solution approaches across different groups of instances -- whiskers show the standard deviation in the groups.}
\label{fig:santos_vs_aco}
\end{figure*}

In Figure \ref{fig:santos_vs_aco}, we assess the quality of our algorithm, in its two versions, by comparing the solutions found by it with the best results reached by the algorithms proposed by Santos and Chagas~\cite{santos2018thief}. For each instance, we consider the best-known solution to be a lower bound on the achievable objective value. Then, we take the average results produced by each approach and compute the ratio between that average and the best objective value found, which gives us the approximation ratio. Note that the higher this metric, the higher the average efficiency of that particular solution method. In the figure, we show the results for the 48 previously defined groups of instances. We report the average approximation ratio obtained for the instances belonging to each group of instances. 

We can see in Figure \ref{fig:santos_vs_aco} that our algorithm has performed significantly better on all groups of instances, especially on those with larger instances. Note that the algorithms proposed by Santos and Chagas~\cite{santos2018thief} are highly affected by the number of items contained in each city, while our framework appears to do better everywhere, and especially well (relatively speaking) on larger instances, where is finds many new best solutions.
Our approaches ACOThOP* and ACOThOP outperform the results achieved in~\cite{santos2018thief} on 419 and 410 out of 432 instances, based on the average solution quality. Regarding the best results found, our approaches have been able to find better solutions for 410 and 402 instances, respectively. On average, considering the best results obtained for all instances, our approaches ACOThOP* and ACOThOP have been, respectively, 320\% and 313\% better than the best solutions found in~\cite{santos2018thief}. In addition, our results show lower standard deviation values, which indicates a better convergence of our algorithm.

To statistically compare the quality of the solutions, we use the Wilcoxon signed-rank test on the results achieved in the 10 independent runs of each solution method. With a significance level of 5\% (\mbox{$p$-value}~$<~0.05$), the performance compared to~\cite{santos2018thief} is as follows on the 432 instances:
\begin{itemize}
    \item ACOThOP* is worse in only 2 cases, there is no difference in 21 cases, and it is better in 409 cases (95\%). 
    \item ACOThOP is worse in only 18 cases, there is no difference in 12 cases, and it is better in 401 cases (93\%).
\end{itemize}

\begin{figure}[!ht]
\centering
\scriptsize
\captionsetup[subfigure]{labelformat=empty, justification=centering}
\centering
 	\setcounter{subfigure}{0}
 	\subfloat[][{\scriptsize Santos and Chagas (2018)
 	
 	Profit = 133925
 	
 	Distance traveled = 54183}] {
 	    \fbox{%
      		\includegraphics[width=0.46\linewidth]{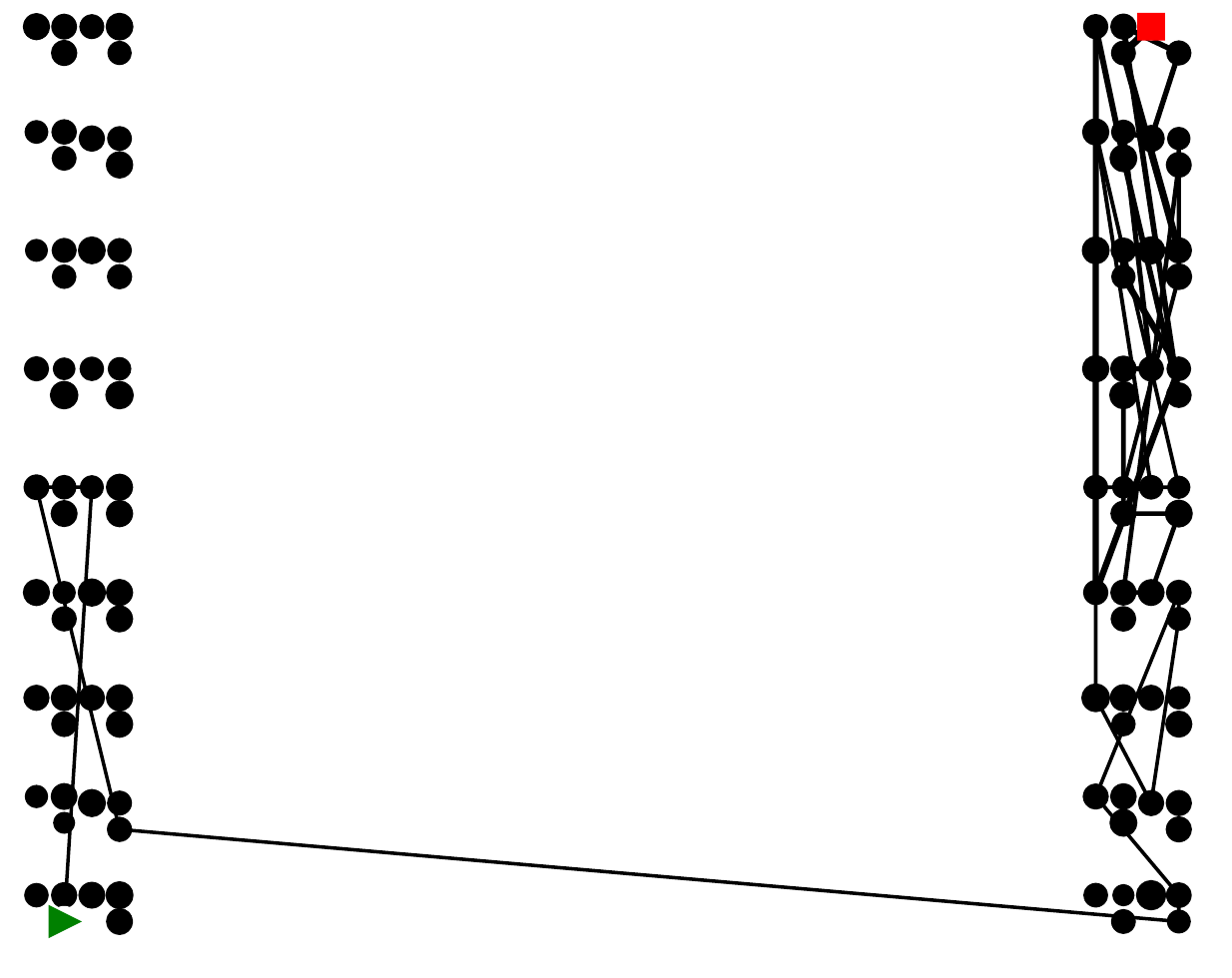}
      	}%
 	}%
 	%
 	%
 	\subfloat[][{\scriptsize ACOThOP*
 	
 	Profit = 474464
 	
 	Distance traveled = 40427}] {
      	\fbox{
      		\includegraphics[width=0.46\linewidth]{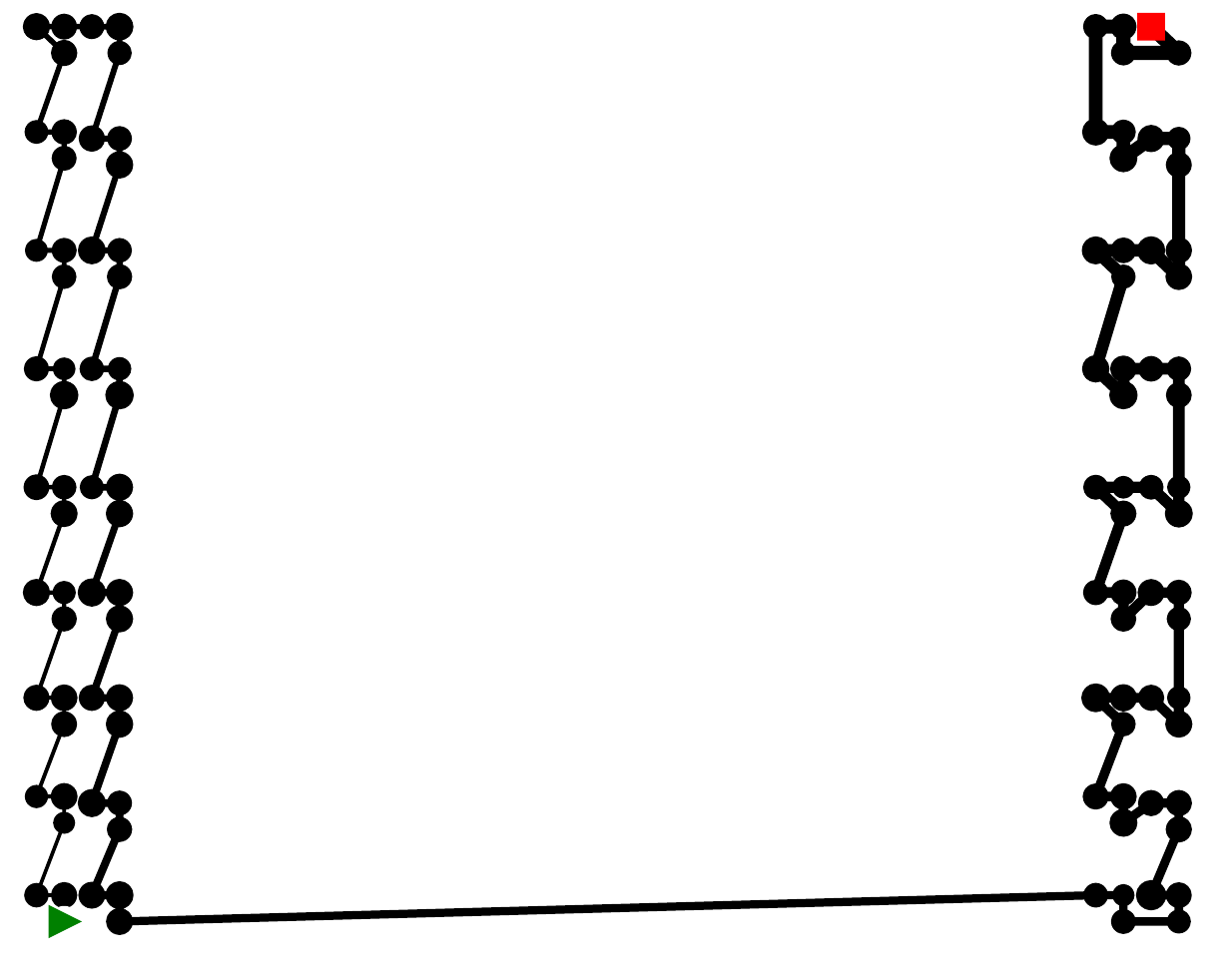}
      	}%
 	}%
\caption{Graphical representation of the best solution found in \cite{santos2018thief} (left) and the best solution found by our approach ACOThOP* (right) for the instance \textit{pr107\_10\_usw\_10\_03.thop}.}
\label{fig:pr107_solutions}
\centering
\end{figure}

Table~\ref{table:solution_structures} summarizes a closer analysis of the solutions found. For each TSP base instance, which resulted in 108 instances each, we show averaged information concerning all the best solutions achieved by each approach. Column $\mathcal{D}$ shows the ratio between the total distance traveled and the number of cities visited by the thief, while columns \textit{\%T} and \textit{\%W} report the percentage spent of the time limit and the percentage used of the knapsack capacity. If values in these last two columns are close to 100\%, then these indicate limiting factors. Furthermore, by comparing the values in column $\mathcal{D}$ from the same TSP base instance, we can see which approach has found the most spread-out routes and/or with more edge crossings. As an example, we show this in  Figure \ref{fig:pr107_solutions} for the instance \textit{pr107\_10\_usw\_10\_03.thop}; this is an instance with a high performance difference between the two shown approaches. The graphical representation of the solutions plots the cities in their respective coordinates. The initial and final cities are represented by a green triangle and a red square, respectively, while black points represent the other cities. The diameter of the point representing a city shows the proportion of profit available in that city. The continuous lines connecting pairs of cities represents the route performed by the thief. The line thickness increases according to the total weight picked by the thief. We can see that our solution has a significantly more efficient route, it travels a shorter distance and, from it, a higher profit has been achieved.

\begin{table}[!ht]
\centering
\footnotesize
\caption{Information on the structure of the best solutions found.}
\setlength{\tabcolsep}{0mm}
\renewcommand{\arraystretch}{0.9}
\begin{tabular*}{\hsize}{@{}@{\extracolsep{\fill}}llrrrlrrrlrrr@{}}
\toprule
\multirow{2}{*}{TSP base} &  & \multicolumn{ 3}{c}{Santos and } &  & \multicolumn{ 3}{c}{\multirow{2}{*}{ACOThOP*}} &  & \multicolumn{ 3}{c}{\multirow{2}{*}{ACOThOP}} \\ 
\multicolumn{1}{c}{\multirow{2}{*}{(\tt XXX)}}  &  & \multicolumn{ 3}{c}{Chagas (2018)} &  & \multicolumn{ 3}{c}{} &  & \multicolumn{ 3}{c}{} \\
\cmidrule{3-5} \cmidrule{7-9} \cmidrule{11-13}
 &  & \multicolumn{1}{c}{$\mathcal{D}$} & \multicolumn{1}{c}{\%T} & \multicolumn{1}{c}{\%W} & \multicolumn{1}{c}{} & \multicolumn{1}{c}{$\mathcal{D}$} & \multicolumn{1}{c}{\%T} & \multicolumn{1}{c}{\%W} &  & \multicolumn{1}{c}{$\mathcal{D}$} & \multicolumn{1}{c}{\%T} & \multicolumn{1}{c}{\%W} \\ 
\midrule
eil51 &  & 12.6 & 92.9 & 73.5 &  & 9.7 & 93.1 & 82.6 &  & 9.8 & 93.0 & 81.9 \\ 
pr107 &  & 925.2 & 96.0 & 64.7 &  & 537.3 & 99.7 & 81.8 &  & 543.8 & 99.7 & 81.0 \\ 
a280 &  & 35.3 & 97.9 & 50.5 &  & 13.2 & 98.9 & 81.4 &  & 13.7 & 98.9 & 80.3 \\ 
dsj1000 &  & 178520.2 & 98.6 & 33.7 &  & 30290.7 & 93.5 & 82.1 &  & 31204.3 & 93.6 & 81.1 \\ 
\bottomrule
\end{tabular*}
\label{table:solution_structures}
\centering
\end{table}

We can observe in Table~\ref{table:solution_structures} that the routes found by our ACO algorithm, in its two versions, are more efficient than those found by Santos and Chagas~\cite{santos2018thief}. Also, we note that the ratio between the total distance traveled and the number of cities visited is higher for the best solutions found in~\cite{santos2018thief}, especially for instances with more cities. This behavior directly impacts solutions because they can be quickly limited by the travel time limit, which can be seen when analyzing the columns referring to the solutions found in~\cite{santos2018thief}. As our ACO algorithm -- together with our packing routine that fills the knapsack more -- has been able to find more efficient routes, a better balance between the limiting factors has been obtained, which resulted in significantly better solutions (see again Figure~\ref{fig:santos_vs_aco}).

\section{Concluding remarks}
\label{sec:conclusions}

In this work, we have approached the Thief Orienteering Problem (ThOP), a recent academic multi-component problem that combines two classic combinatorial optimization problems: the Orienteering Problem and the Knapsack Problem. 
We have proposed a two-phase heuristic algorithm based on Ant Colony Optimization, and we have studied the effect of the components using automated algorithm configuration.
Our experiments have shown that the best configurations as well as the average configuration are better on over 90\% of the 432 instances with an average fitness improvement higher of over 300\%; the largest improvements are on the largest instances, when compared to the best solutions in the literature. 
Based on our analysis, this is due to the efficiency of the ant colony optimization used to determine the thief's route together with our novel, randomized packing routine. 

As future work, we will investigate exact algorithms to solve small and mid-sized ThOP instances to establish global optima. Another interesting study will be to address a version of the problem that considers multiple thieves in order to provide a more generic problem, for example, to take a more fundamental approach to the above-mentioned scenarios of the politicians campaigning or the rescue-teams checking safety places.

\vspace{2mm}\noindent\textbf{Acknowledgments.}
This study has been financed in part by Coordena\c{c}\~{a}o de A\-per\-fei\-\c{c}o\-a\-men\-to de Pessoal de N\'{i}vel Superior - Brazil (CAPES) - Finance code 001. The authors would also like to thank Funda\c{c}\~{a}o de Amparo \`{a} Pesquisa do Estado de Minas Gerais (FAPEMIG), Conselho Nacional de Desenvolvimento Cient\'{i}fico e Tecnol\'{o}gico (CNPq), Universidade Federal de Ouro Preto (UFOP) and Universidade Federal de Vi\c{c}osa (UFV) for supporting this research.

%
%

\bibliographystyle{splncs04}
\bibliography{main}
\end{document}